\documentclass[letterpaper, 10 pt, conference]{ieeeconf}  

\IEEEoverridecommandlockouts                              

\usepackage{cite}
\usepackage{amsmath,amssymb,amsfonts}
\usepackage{algorithmic}
\usepackage{graphicx}
\usepackage{textcomp}
\usepackage{xcolor}
\usepackage{subfigure}
\title{\LARGE \bf Hybrid Classical/RL Local Planner for Ground Robot Navigation}

\author{Vishnu D. Sharma, Jeongran Lee, Matthew Andrews, Ilija Had\v{z}i\'{c}
\thanks{Work done while V. D. Sharma was a summer intern from the University of Maryland, College Park, USA. }
\\
Nokia Bell Labs - Murray Hill, USA
}
\graphicspath{{./figs/}}

\newcommand{\cA}{{\cal A}}
\newcommand{\cS}{{\cal S}}

\begin{document}

\maketitle

\begin{abstract}
  Local planning is an optimization process within a mobile robot navigation stack that searches for the best velocity vector, given the robot and environment state. Depending on how the optimization criteria and constraints are defined, some planners may be better than others in specific situations. We consider two conceptually different planners. The first planner explores the velocity space in real-time and has superior path-tracking and motion smoothness performance. The second planner was trained using reinforcement learning methods to produce the best velocity based on its training ``experience''. It is better at avoiding dynamic obstacles but at the expense of motion smoothness. We propose a simple yet effective meta-reasoning approach that takes advantage of both approaches by switching between planners based on the surroundings. We demonstrate the superiority of our hybrid planner, both qualitatively and quantitatively, over the individual planners on a live robot in different scenarios,  achieving an improvement of \textbf{26\%} in the navigation time.
\end{abstract}

\section{Introduction}
A mobile robot navigation stack is broadly responsible for safely (and desirably optimally) getting the robot from its present position to the goal while respecting externally or internally imposed constraints.
Components of a path- and motion-planning and control subsystem can be broadly categorized into global planners, local planners/controllers, and motion controllers, which are typically deployed in concert. Global planner finds the path toward the goal location, often expressed as a set of waypoints that the robot must visit. The local planners are responsbile for generating the velocity vectors to lead the robot towards the next waypoint.

In a known map, global planners are optimal as they utilize the global costmap, but are brittle in the presence of unknown (and discovered after the fact) dynamic obstacles, such as humans, clutter, unmapped fixtures, and other vehicles. Local planners, on the other hand, can react well in such situations. Additionally, local planners take less time to compute and thus process the data at a higher frequency.

Local planning in velocity space can be characterized as an optimization process (which may in practice produce suboptimal, but acceptable solutions), whose optimization criteria include distance to the next waypoint (or the goal), clearance around the obstacle, smoothness of motion, energy efficiency, and the like. For this discussion, we broadly classify the implementations into classical and learning-based approaches. Classical planners explore the velocity space and evaluate each proposed velocity against the constraints and the optimization criteria in real-time. To find an optimal solution a classical planner must often search the entire space of admissible velocities, which, depending on the size of the planning window, the number of degrees of freedom, and the complexity of constraints, can be a challenging process. 

Learning-based planners are exposed to various situations offline and trained to map the robot state to the deemed best velocity, typically using a neural network. The complexity of searching and evaluating the solution is moved to an offline training process. The real-time computation becomes the model inference and does not involve explicit search. The performance of these planners strongly depends on how the training environment was set up, the variety of situations the robot has been exposed to, and how well the dynamics of the robot were captured during the training. Typically, reinforcement learning (RL) techniques are used here. As with all learning-based algorithms, false results are possible and it is impossible to guarantee that the RL-planner will always produce optimal or even correct solutions. Nevertheless, RL-planners have been shown to produce useful results that generalize well\cite{guldenring2020learning, van2017extended, liu2020robot, KastnerZBLSLM21, patel2021dwa, nakhleh2023sacplanner}. In our previous work\cite{nakhleh2023sacplanner}, we designed an RL-planner with superior obstacle-avoidance performance compared to a widely used Dynamic Window Approach (DWA) planner\cite{fox1997dynamic}, but the price of this improvement was an uneven and jerky motion, even when no dynamic obstacles were present in the robot path.

This lack of smoothness limits the attractiveness of RL-based local planners as a general solution. If the robot is moving through a large open space, or if it is moving in a maze-like structure with known, fixed walls, it can stay close to the global plan. Classical local planners typically excel at generating smooth motion toward the goal. In this case, the instantaneous decisions of an RL-based planner are overkill and can lead to rapid changes in velocity that do not provide any benefit. Although reworking the training process to penalize uneven motion may lead to improved behavior, it is unclear how the two opposing criteria would reflect on general performance. Further, conceiving a new training process and designing an improved reward function is an arduous effort that is often subject to trial and error.

Alternatively, one can simply recognize that an RL-based planner performs better when confronted with an unexpected or dynamic obstacle, whereas a classical planner performs better when the robot simply needs to track the global plan. In this context, a pragmatic solution is to conceive a decision tree that recognizes the current situation and switches to using the planner known to produce a better solution. The existing works~\cite{dey2023learning, raj2023targeted} have proposed learning the switching criteria with a neural network, which requires further training and may suffer from the typical shortcomings of teh learning-based approaches, such as generelizability.

In this paper, we propose a simple hybrid planner that detects if the global plan is obstructed by an unexpected obstacle and picks the solution provided by a (more responsive) RL-planner. Otherwise, it takes the solution provided by a classical planner. We demonstrate via experiments that this hybrid approach responds well in the obstructed case while maintaining smooth performance in the non-obstructed case.

\section{Related Work}
Local planners play an important role in obstacle avoidance and have been a topic of interest for a long time~\cite{sanchez2021path}.
Classical planning approaches, which do not employ learning, are widely used across robotics applications. Reactive replanning~\cite{fox1997dynamic,rosmann2017kinodynamic}, artificial potential field~\cite{bin2011research}, and fuzzy logic-based approaches~\cite{yan2016mobile} are examples. One such widely used method, proposed by Fox et al.~\cite{fox1997dynamic} and called Dynamic Window Approach (DWA) planner, uses reactive replanning and has been frequently used as the baseline planner by the Robot Operating System (ROS) navigation stack\cite{rosnavigation}. Because of its widespread use and availability in open-source community, ROS implementation of DWA has often been used as the baseline, despite the algorithm being relatively old. For this reason, we baseline our results to DWA.

An alternate way to design a local planner is to learn the system model using data and fine-tune the learning model in a new environment. Such learning-based approaches have been introduced in the past few years and have been growing rapidly in number. A deep reinforcement learning (DRL) framework is often used for training in such approaches as it allows the robot to interact with the environment without needing data collection and annotation~\cite{van2017extended, Gldenring2019ApplyingDR,guldenring2020learning,liu2020robot,KastnerZBLSLM21,patel2021dwa,nakhleh2023sacplanner}. The framework proposed by G{\"u}ldenring et al.~\cite{Gldenring2019ApplyingDR}, which uses 2D local map and waypoints from the global plan for state representation, was used as the base for the development of many subsequent works. In our previous work\cite{nakhleh2023sacplanner}, we studied and compared classical planners and different RL network architectures and proposed a method that used a polar representation of the costmap in state representations. This network, named SACPlanner, was trained in a simulation environment and tested on live robots. SACPlanner outperformed other approaches, including DWA, in safely avoiding collisions with static and dynamic obstacles. The practical result was a more responsive planner, but slower and jerky motion caused by the robot trying to move cautiously even when the path ahead was clear. Arguably, this behavior can be improved with training in a higher-fidelity simulation environment, but at the risk of breaking other
desirable properties achieved during the original training.

One way to get the benefits of different types of planners is to use an ensemble of methods with user-defined control. The use of such \textit{hybrid} planning strategies to harness both classical and learning-based approaches is a fairly recent development~\cite{von2020combining}. Existing work in the literature has explored both hybrid robotic planners consisting of classical approaches~\cite{orozco2019hybrid} and planners using learning-based approaches~\cite{lu2020hybrid}. Existing hybrid planners combining classical and learning-based approaches lie in the middle of this spectrum and aim to combine the model-based classical approaches and data-based learning approaches by switching between them.

Almadhoun et al.~\cite{almadhoun2021multi} use heuristics-based criteria to switch between a  classical and a learning-based approach to generate viewpoints for 3D reconstruction. Linh et al.~\cite{linh2022all} and Dey et al.~\cite{dey2023learning} study ground robot navigation but they rely on neural networks for learning and focus on high-level planning. Raj et al.~\cite{raj2023targeted} also proposed a neural network-based switch, but they focused on social navigation only. In contrast, our work contributes towards the development of a local planner that uses a hybrid approach that combines classical and learning-based methods. We design a heuristics-based logic for switching between a DWA planner and SACPlanner, enjoying the benefits of both. This hybrid planner exhibits a superior performance with a simple design which forgoes the need to train another neural network for switching.

\section{Preliminaries}
\label{sec:background}
The local planner/controller is responsible for generating the velocity vector that makes progress toward the goal or the next waypoint. Some implementations explore the velocity space and score candidate velocities based on forward simulation in the configuration space (which, strictly speaking, makes them planners), whereas others solve a constrained optimization problem that maps the state to an action (which, strictly speaking, makes them controllers). These planners/controllers can either generate the motion in the velocity space and leave it to a lower-level motion controller to generate the actuation or directly solve for actuation. A motion controller (if present separately from the local planner/controller) generates the actuation that delivers the desired velocity vector. In this paper, we focus on local planning/control in velocity space and for simplicity we use the term ``local planner" to mean any subsystem that generates the desired velocity vector based on the present robot configuration (specifically, the robot pose) and the state of the surrounding environment (specifically, the next waypoint pose, goal pose, and perception of obstacles). In the following subsections, we describe the classical and the learning-based local planners used in our work, DWA and SACPlanner respectively. 

\subsection{Dynamic Window Approach (DWA)}
\label{subsec:dwa}
DWA planner generates a set of admissible velocities, which are the velocities that can be reached given the present velocity and the robot dynamic constraints (i.e., acceleration limits). For each admissible velocity, DWA performs a forward simulation to calculate the resulting trajectory should the robot use this velocity. Finally, each simulated trajectory is scored and the one with the lowest cost is selected. The objective function reflects
progress towards the goal, clearance from the obstacle, adhering to the global plan (distance to the waypoint), and twirling.

DWA considers the robot's dynamics and the overall motion is a series of arcs determined by the angular and linear velocity, where each planning step produces one such arc. If there are no obstacles in the path, the planner will pick the arc that best advances the robot toward the next waypoint, as the distance from the global plan is part of the cost function. In an obstacle-free environment, the selection of the best velocity will be the balance between sticking to the global plan (advancing to the nearest waypoint) and advancing towards the global (cutting corners in the global plan to reach the goal sooner). Parameters allow the user to tune the planner to balance between one behavior and the other. While this single-arc planning is works well in general, situations, as described below, may need complex velocity profile is needed, making DWA ineffective in the scenarios. 

If there is an obstacle in the path, the obstacle-distance component of the cost function will start to dominate and the arcs that point away from the global path will have a lower cost, consequently making the robot deviate from the global plan or the goal.
As the robot steers away, the plan-distance and goal-distance components of the cost function will equalize and the robot will gravitate back to the plan. Three cases are possible next: 1) the robot may have made sufficient forward progress that the next waypoint is behind the obstacle, in which case the local planner will return the robot to the path determined by the global plan; 2) the robot may turn back into the obstacle make a motion towards it, and steer away from the obstacle again, but this time being in a more difficult situation due to obstacle proximity; 3) the global planner may trigger and generate a new set of waypoints that will guide the robot around the obstacle.

Ideal local planners should always result in the first case, which would make it able to deal with obstacles on its own.  The second case can often lead to a live-lock that manifests itself by a robot approaching the obstacle and indecisively oscillating without making progress. In some cases, the collision may occur due to sensor limitations. Namely, in our experiments, we saw collision because the LiDAR sensor that we used has the minimum-range distance. Once the robot gets too close to the obstacle, the reflections are not registered and the robot charges into the obstacle. Augmenting the robot with the second, short-distance sensor to prevent these collisions resulted in described live-locks.

\begin{figure}[ht!]
\vspace{3mm}
  \centering
  \includegraphics[height=0.15\textwidth]{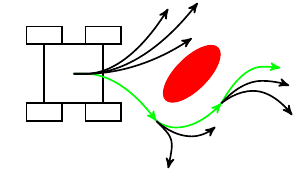}
  \caption{Confronting an obstacle in a series of arc-motions.}
  \label{fig:arcmotion} 
\end{figure}

We argue that these shortcomings are direct consequences of single-arc
motion planning that DWA uses. Successful obstacle avoidance requires three consecutive arcs as shown in green in Fig.~\ref{fig:arcmotion}. The first arc pushes the robot away from the obstacle, the second sends it back on track once the obstacle has been successfully navigated around, and the third realigns the direction to the plan. DWA planner simply does not explore the space beyond one velocity vector and longer simulation time simply extends the arc into the space that is not relevant for evaluating the motion. We confirmed this with a series of experiments, tuning one parameter at a time while tracing the DWA code to find the root cause. All tests pointed to the lack of visibility into the subsequent arcs that may follow the one being scored.
\vspace{-1mm}

\vspace{-1mm}

Extending the planner to explore a series of velocity vectors scales exponentially with the number of composite arcs to explore. The sequence of arcs shown in green in Fig.~\ref{fig:arcmotion} successfully navigates around the obstacle, but to select it, all three arcs in the sequence must be scored. At each step, there are multiple candidates (arcs shown in black) that must also be scored to find the optimal path around the obstacle.

The third case is commonly used in practice to counter the above-described problem. Replanning on the global level is achieved either by running the global planner periodically at low rate (e.g., once every few seconds) or by having ``patience" timers built into the navigation stack that trigger the global planner when deemed necessary.
Careful tuning of cost-function weights, timer values, and other constraints, can result in satisfactory and safe performance of the navigation stack, but this process is arduous and practitioners often resort to trial and error.

More cases the local planner can deal with when left on its own, more robust navigation stack will be when the assistance from the global planner is turned on. In our evaluation, we disallow global replanning, because we are interested in performance of the local planner alone, rather than the whole navigation stack. This results collision-avoidance performance that some practitioners may find surprisingly poor, but this is due to confusing the performance of the whole navigation stack as opposed to the performance of the local planner alone.

\subsection{SACPlanner}
\label{subsec:sacplanner}
SACPlanner, which we previously developed\cite{nakhleh2023sacplanner}, is a RL-based planner that outperforms DWA in challenging situations. We have experimentally shown that it successfully resolves the problem described in Section~\ref{subsec:dwa}. An intuitive explanation is that the arc motion it selects is statistically the most likely to be the correct first step in the chain of velocity vectors that will avoid the obstacle and put the robot back on the planned path. There is no velocity-space exploration and although a single compute step is more complex, it eliminates the problem of exponential scaling.

SACPlanner uses a polar representation of the local costmap as the input to the neural network (see Fig.~\ref{fig:wp_polar}) and outputs an angular and linear velocity pair as the action for the robot. It uses the Soft Actor-Critic~\cite{haarnoja2018soft} method for training with a mixture of dense and sparse rewards that quantify the robot's progress towards the goal and collision-avoidance, similar to DWA's objective function. Even though it is trained in a simulation environment, we have shown that it generalizes well\cite{nakhleh2023sacplanner} and using  polar representation of the local costmap as the state helps in sim-to-real transfer without fine-tuning. We demonstrated that a real robot can successfully execute PointGoal navigation in complex mazes and with unexpected obstacles, whereas DWA typically ends up in a state from which it does not make meaningful progress toward the goal or in some cases collides. We have experimentally determined that when the collision occurs, it is typically due to the sensor limitation. Namely, the LiDAR we use has the minimum range below which it becomes ``blind''. Whenever the collision occurred, it would be because the DWA planner pushed the robot too close to the obstacle to provoke the sensing problem. We believe that if the sensing were augmented to resolve this problem the problem would simply morph into stalling the robot in front of the obstacle. SACPlanner, on the other hand, never brought the robot into such a situation and successfully avoided the obstacles despite the sensing limitation.

A learning-based planner effectively retains the mapping between the input and the output as network weights. This results in SACPlanner potentially learning how to behave in a conservative fashion to safely avoid obstacles. Whereas, DWA is limited to executing motion on circular arcs, SACPlanner can traverse complex trajectories. However, as SACPlanner looks at the costmap in an instant only, the robot's motion is jerky and it moves at a slower speed, making it inefficient even when there is no obstacle ahead. 

These two planners represent two seemingly contrasting planning approaches. Choosing one planner from these is essentially a tradeoff between \textit{smoothness} and \textit{responsiveness}. While DWA is more suitable for moving on a static map, SACPlanner is better equipped for successful navigation in complex and dynamic environment. We use this idea to propose a hybrid approach that uses both planners for safer and more efficient planning.

\section{Hybrid Local Planner}
\label{sec:approach}
\begin{figure}
    \centering
    \vspace{2mm}
    \includegraphics[width=0.98\linewidth]{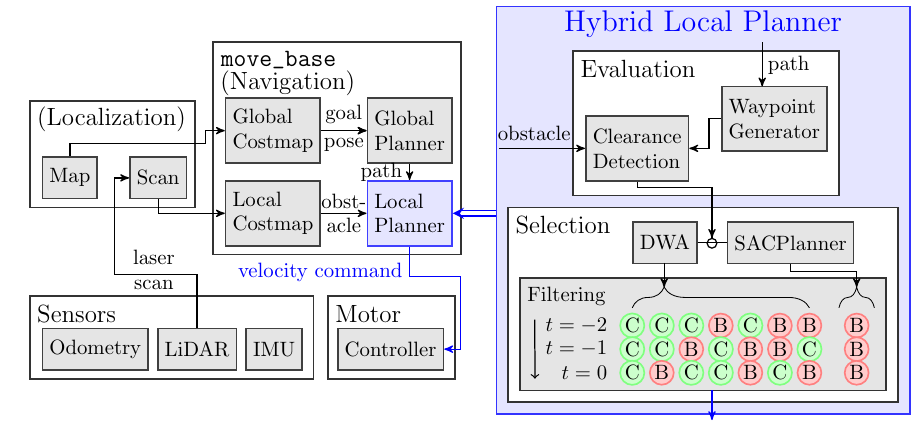}
    \caption{ROS framework and the architecture of our hybrid local planner.}
    \label{fig:ros}\vspace{-3mm}
\end{figure}

We propose a hybrid local planning approach that combines the benefits of a classical planner and a learning-based planner. Specifically, we run DWA and SACPlanner in parallel and switch between them based on the clearance ahead of the robots. Fig.~\ref{fig:ros} shows the architecture of our implementation. The box labeled {\tt move\_base} comes from standard ROS navigation stack and we modified the local planner plugin to include the DWA code verbatim from the ROS navigation stack, our SACPlanner implementation, along with the code that implements the switching policy. This is illustrated by the box on the right labeled Hybrid Local Planner.

\subsection{Waypoint Generation}
\label{subsec:waypoint}
\begin{figure}[!h]
    \centering\vspace{-4mm}
    \includegraphics[width=0.7\linewidth]{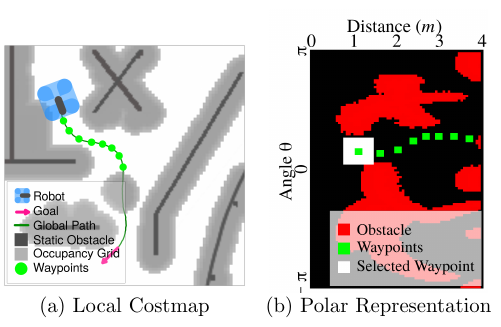}\vspace{-2mm}
    \caption{Waypoint generation.}\vspace{-4mm}
    \label{fig:wp_polar}
\end{figure}

First, we use the method proposed by G\"{u}ldenring et al.~\cite{Gldenring2019ApplyingDR} to find waypoints on the local map. We use the waypoints both to decide which local planner to use and also to create the goal in case the SACPlanner is selected. To generate the waypoints, the global plan leading to the goal, generated with Dijkstra's algorithm, is downsampled and a fixed number of waypoints, 8 in our case, are selected on the local costmap, as shown in Fig.~\ref{fig:wp_polar}(a).  This set of waypoints helps the robot align with the global plan and thus also avoids local minima. The first waypoint not on the obstacles is fed to SACPlanner as the goal in the polar image as Fig.~\ref{fig:wp_polar}(b).

\subsection{Clearance Detection}
\label{subsec:linearization}
\begin{figure}[!h]
\vspace{1mm}
    \centering
    \includegraphics[width=\linewidth]{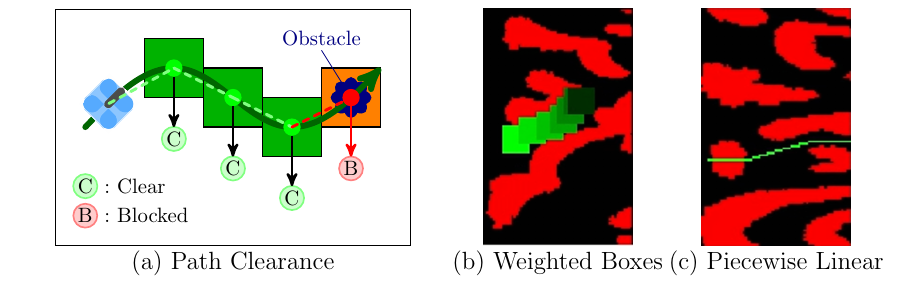}\vspace{-2mm}
    \caption{Clearance detection.}
    \label{fig:wpc}
\end{figure}

To switch between the planners, the robot needs to determine the clearance ahead. In order to enable an early response, we find if the path without any dynamic obstacle can be traversed without collision. We use the waypoints generated on the local costmap for clearance detection. We check if this path is obstructed anywhere on the local map. If the whole path is unobstructed, we consider the path to be \textit{clear}. Otherwise, this path is considered as \textit{blocked}. Fig.~\ref{fig:wpc}(a) demonstrates this approach. The clearance detector can be defined as weighted boxes around the waypoints as in Fig.~\ref{fig:wpc}(b). But the size of the box should be tuned since a smaller box can miss obstacles residing in between gaps, whereas a bigger box cannot get through a narrow pathway smoothly. To avoid this, we create a piecewise linear trajectory to approximate the path the robot would have followed if there were no dynamic obstacles in the environment. The path shown in the example Fig.~\ref{fig:wpc}(c) is detected as not clear since the initial part of the trajectory is blocked by an obstacle (shown as a red blob).\vspace{-2mm}

\subsection{Filtering}\vspace{-1mm}
\label{subsec:filtering}
Noise in the sensor data could result in the clearance detector rapidly flip-flopping between the two planners if only the latest clearance is used for planner selection. For stabilization, the switch should take place only when we are confident about the presence of an obstacle on the path. The typical way to tackle noise in such a situation is to check the likelihood $\mathcal{L}(b | O_{t-n:t})$ of the path being blocked $b$ based on the past $n$ observations till the current time $t$. If the likelihood of obstacles is higher than a user-defined threshold $\tau$, we consider the path to be blocked.

We implement this strategy as a filter that keeps track of the last $n=3$ path clearance statuses from the detector. If all the statues indicate a blocked path, we use the SACPlanner, effectively using $\tau = 1$. Otherwise, DWA is used. This scheme is visualized in the \textit{Filtering} step (right bottom box) in Fig.~\ref{fig:ros}. This design helps in using the SACPlanner when the sensors strongly indicate the presence of an obstacle on the path and results in efficient navigation as the comparatively smoother and faster approach, DWA, is used most of the time and the switching occurs only if necessary. 

We use Robot Operating System (ROS) to implement this pipeline in C++ and Python. Our approach runs DWA and SACPlanner in parallel and switches between them by using the velocity prescribed by the selected planner. 

\section{Implementation Details}\vspace{-1mm}
\subsection{SACPlanner}
We now provide some more details about the SACPlanner. See \cite{nakhleh2023sacplanner} for the full description. SACPlanner is a Reinforcement Learning (RL) based planner with a state space $\cS$, an action space $\cA$, and a reward function $R(\cdot,\cdot)$. 
The actions are simply the linear/angular velocity pairs $(v,\omega)$. 
For the state space, we use an image representation that allows the RL machinery that has been developed for video games. Specifically, the RL state is an image that combines a goal point and all the obstacles that are either derived from the static map or sensed by LiDAR. 

The goal point is one of the waypoints already discussed in Section~\ref{sec:approach}. In particular, we select the first waypoint that does not coincide with an obstacle. We combine this waypoint with the Occupancy Grid representation of the ROS costmap (that represents the nearby obstacles). We then create a polar representation of the waypoint and obstacles, where the horizontal axis represents the distance from the robot and the vertical axis represents the angle. (See Fig.~\ref{fig:wp_polar}(b) for an example.)

\begin{figure}[h]
	\centering
	\subfigure{\label{fig:dummy_cart}
	\includegraphics[height=0.12\textwidth]{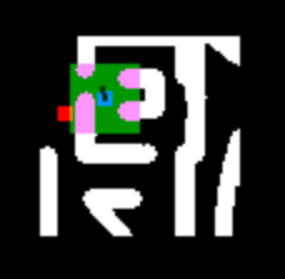}}
	\subfigure{\label{fig:dummy_polar}
	\includegraphics[height=0.12\textwidth]{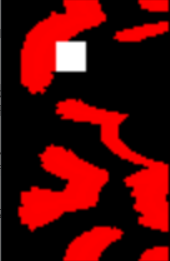}}
	\vspace{-2mm}
	\caption{Dummy training environment (left) and the associated polar costmap (right).}\label{fig:dummy}\vspace{-1mm}
\end{figure}

To train SACPlanner it is convenient and faster to train it offline than in real time, and so we utilize a simulated ``dummy environment''. For each training episode, we pick a synthetic obstacle map and place a robot starting point and a waypoint as in Fig.~\ref{fig:dummy} (left). The episode is successful if the robot reaches the waypoint. The RL state during the training is the associated polar costmap, as described above and shown in Fig.~\ref{fig:dummy} (right). 

We train SACPlanner using a Soft Actor-Critic (SAC) approach~\cite{haarnoja2018softa, haarnoja2018softb}, where the actor is a policy function and the critic evaluates the actor-value function. SAC augments the standard RL objective with an additional entropy maximization term. We also use the RAD~\cite{laskin2020reinforcement} and DrQ~\cite{kostrikov2020image} methods that apply a variety of image augmentations when training the actor/critic functions. 

The reward function $R(s,a)$ for taking action $a$ in state $s$ is defined as follows. Let $(d_{\mathrm{old}}, \theta_{\mathrm{old}})$ be the distance and bearing
to the next waypoint in state $s$, let $s'$ be the new state after taking action $a$, 
and let $(d_{\mathrm{new}}$, $\theta_{\mathrm{new}})$
be the distance and bearing in state $s'$. 
$$
\begin{array}{rl}
    & R(s,a)=\left(d_{\mathrm{old}}- d_{\mathrm{new}}\right) \cdot\left(1 \mbox{~if } d_{\mathrm{old}}- d_{\mathrm{new}} \ge 0, \mbox{~else } 2 \right) \\
    &+\left(|\theta_{\mathrm{old}}|-|\theta_{\mathrm{new}}|\right)\cdot\left(1 \mbox{~if } |\theta_{\mathrm{old}}|-|\theta_{\mathrm{new}}| \ge 0, \mbox{~else } 2 \right) \\
    &-R_\mathrm{max} \cdot\left( 1 \mbox{~if collision, else } 0 \right)  \\
    &+R_\mathrm{max} \cdot\left( 1 \mbox{~if } d_{\mathrm{new}}=0, \mbox{~else } 0\right)\\&-G(s'),
  \label{eq:rew}
\end{array}
$$
where $R_\mathrm{max}$ is the reward/penalty for reaching the waypoint or hitting an obstacle, and $G(s')$ is the product of a truncated Gaussian kernel centered at the robot location and the occupancy grid in state $s'$. (The kernel is represented by the green square in Fig.~\ref{fig:dummy}.) We incentivize direct navigation by doubling the penalty for moving away from the waypoint vs.\ moving towards it.
After 10000 training episodes, we achieve a 98\% episode success rate. For more details on the training performance see \cite{nakhleh2023sacplanner}.

\subsection{Running the planners in parallel}
To implement the hybrid planner we used the \texttt{move\_base} ROS package \cite{rosmovebasewiki}. We instantiate three planners using its base planner class: (1) DWA, (2) SACPlanner, and (3) Hybrid Planner. While the first two compute the appropriate velocity profile, only the latter can send the velocity commands to the motion controller. The hybrid planner calls both DWA and SACPlanner functions for its planning functions, effectively running them in parallel. In the output function, responsible for generating the velocity vector, the planner runs the decision logic described in Section~\ref{subsec:sacplanner} and publishes velocity computed by the selected planner only.

\section{Experiment Setup}
\label{sec:experiemnts}

\begin{figure*}[!ht]
\vspace{1mm}
	\centering
        \vspace{2mm}
 	\subfigure[UNIX maze testbed]{\label{fig:maze}
		\includegraphics[height=0.145\textwidth]{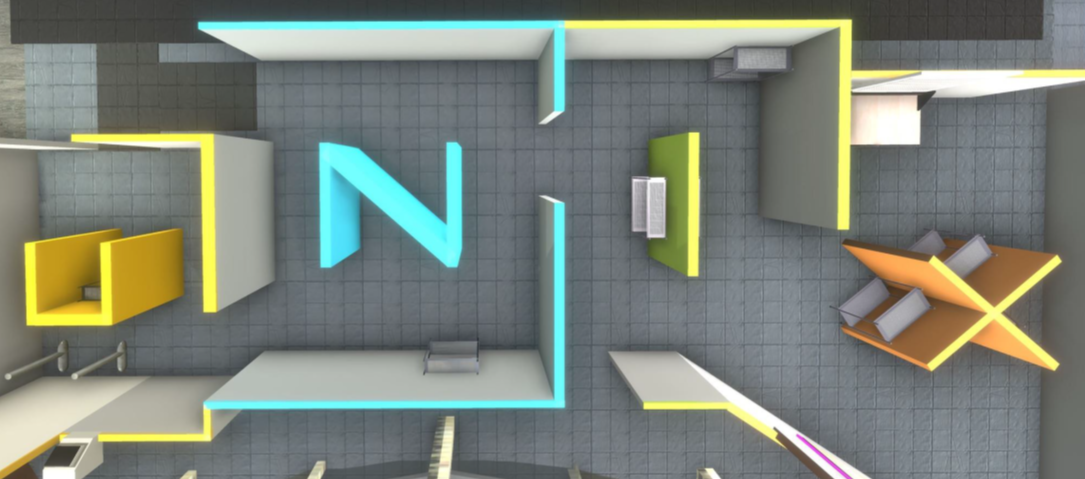}}\hspace{-1mm}
	\subfigure[N-I room doorway]{\label{fig:c1}
		\includegraphics[height=0.145\textwidth]{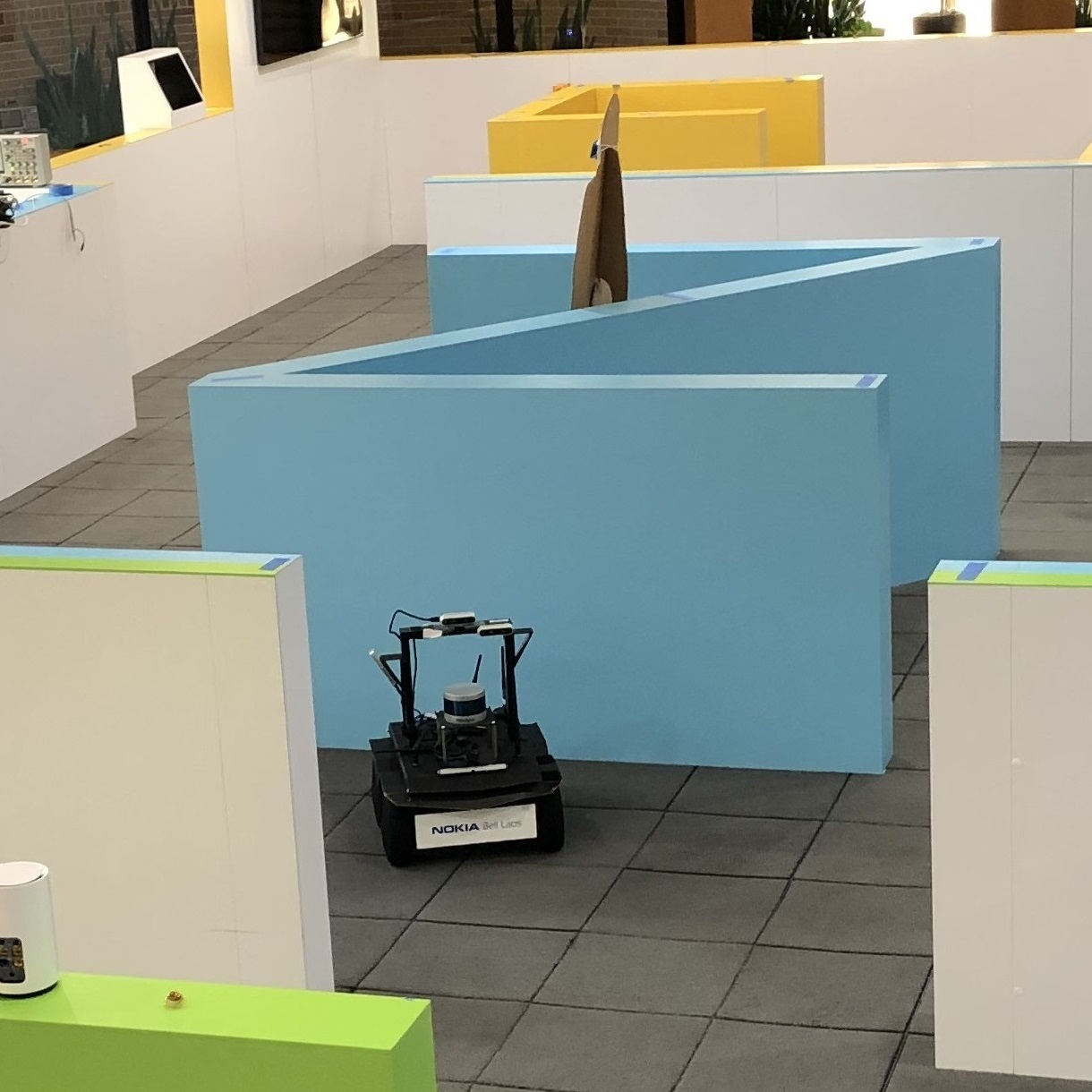}}\hspace{-1mm}
	\subfigure[I-X room cardboard]{\label{fig:c2}
		\includegraphics[height=0.145\textwidth]{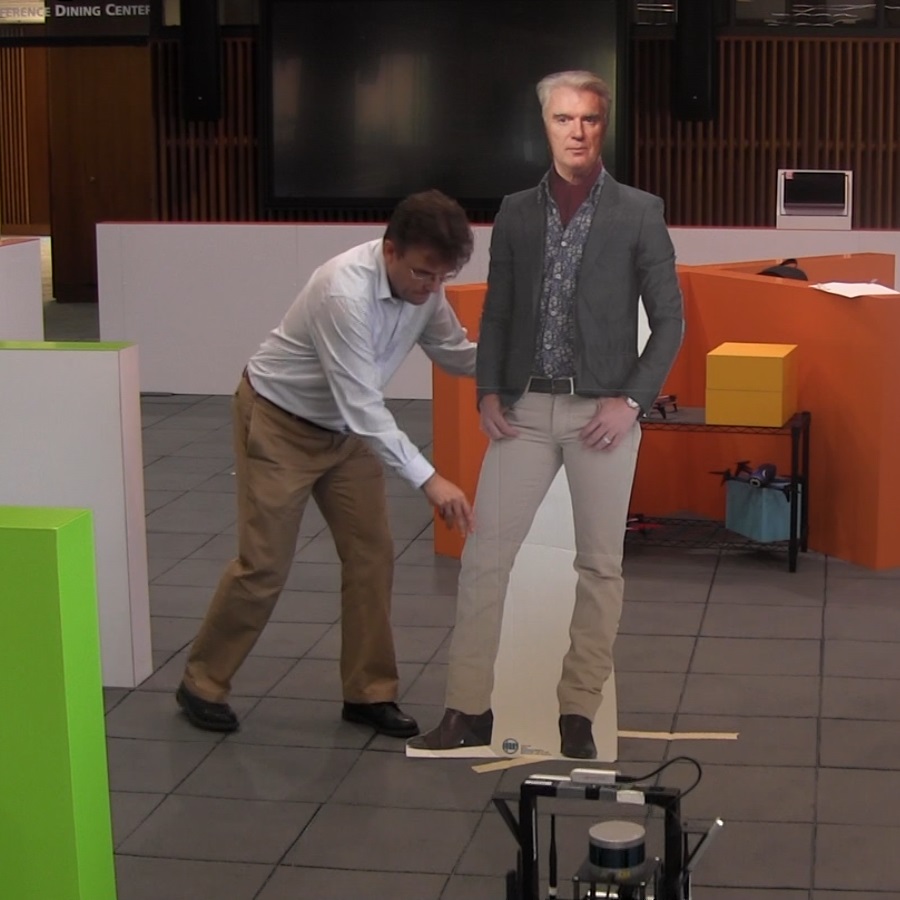}}\hspace{-1mm}	
	\subfigure[Approaching front]{\label{fig:c3}
		\includegraphics[height=0.145\textwidth]{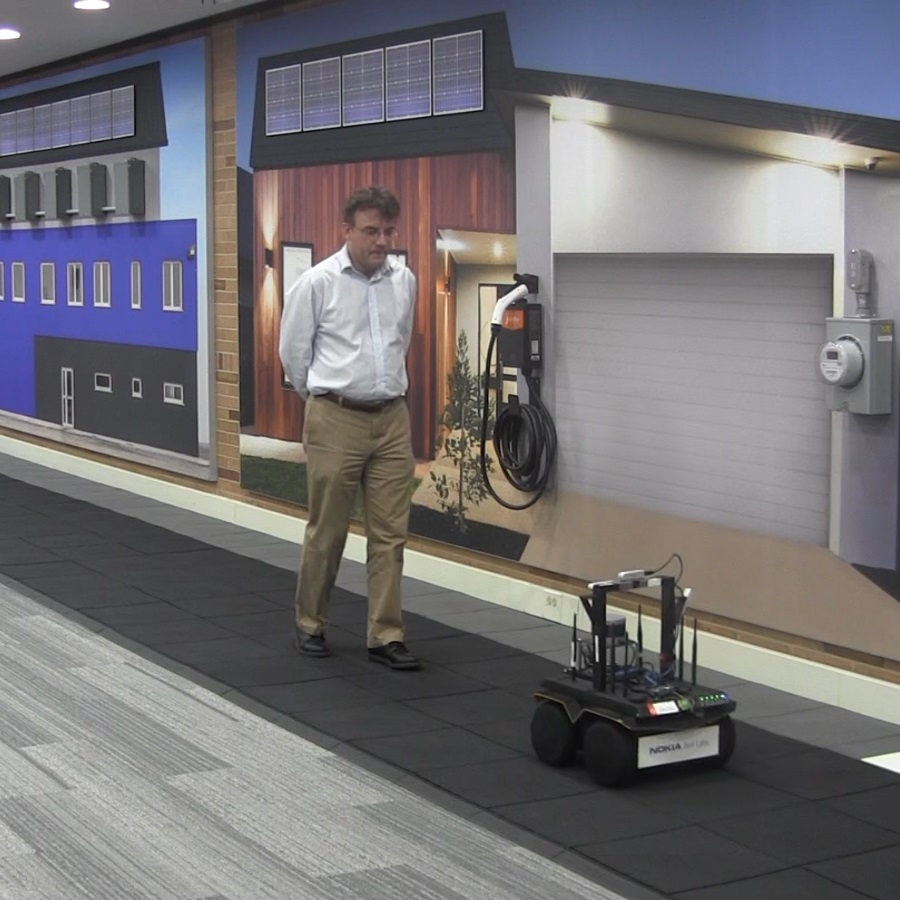}}\hspace{-1mm}	
	\subfigure[Crossing path]{\label{fig:c4}
		\includegraphics[height=0.145\textwidth]{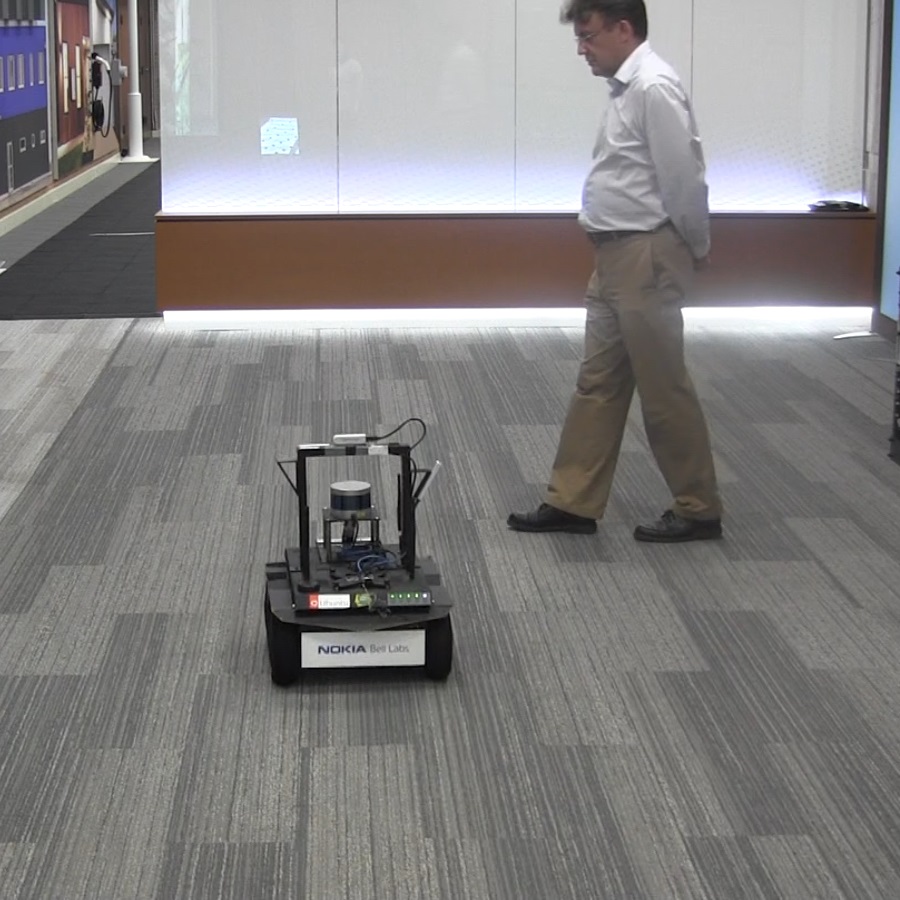}}	
	\vspace{-2mm}
	\caption{Real experimental environment and 4 test case scenarios (C1-4) from left to right.}\label{fig:scenarios}	
\end{figure*}

Our experimental setup is similar to \cite{nakhleh2023sacplanner} for fair comparison. We run experiments using a ClearPath Robotics Jackal robot~\cite{clearpath} in an indoor facility with an open room and a maze with narrow pathways and tight corners, as shown in Fig.~\ref{fig:maze}. We refer to the maze as the \textit{UNIX maze room} after the letters that constitute the walls inside the maze. In the discussion ahead, we also refer to these letters to indicate the location of the experiment. The robot uses a Velodyne LiDAR running at 10Hz for perception and the planner runs at 5Hz (half the sampling rate of the LiDAR). We study different challenging scenarios in a known map as follows:\\\vspace{-2mm}

\textbf{(C1)}
    \textbf{Obstacle-Free Intricate Trajectory:} 
    \label{subsection:c1}
    This task evaluates if the robot is able to traverse on a serpentine trajectory passing through a narrow doorway. Moving on this trajectory requires that the robot make a $180^\circ$ turn. For this setup, we move the robot from Room I to Room N through a narrow doorway as shown in Fig.~\ref{fig:c1}. Successful traversal requires that the robot closely follows the global plan on the known map. The challenge for the local planners lies in adjusting their speed timely while accounting for the inertia to avoid collision with the walls.
    
    \textbf{(C2)}
    \textbf{Unexpected Static Obstacle on Path:} 
    \label{subsection:c2}
    In this case, we test if the robot is able to react well to an unexpected object on the path that appears after the global planning is done and stays at a fixed location for the rest of the experiment. This experiment is realized by moving the robot between Room I and Room X, as shown in Fig.~\ref{fig:c2}. Here we use a life-sized cutout of a person as the static obstacle and place it on the robot's global path after the robot starts moving. This setup is similar to \textit{Doorway} setting in Raj et al.~\cite{raj2023targeted}. Successful execution requires that the robot moves past the obstacle from the side.
    
    \textbf{(C3)} 
    \textbf{Dynamic Obstacle on Path:}
    \label{subsection:c3}
    Here we test the robot's ability to dynamic obstacles on the robot's global path. For this, we move the robot in a straight line in an open area and a pedestrian walks quickly toward the robot after the robot starts moving on the global path, in a straight line. An obstacle moving at a high speed makes it difficult for the local planner to react in time as the obstacle only shows up after it has entered the robot's local map and keeps changing the location. This situation is shown in Fig.~\ref{fig:c3} and is similar to the \textit{Frontal} setting in Raj et al.~\cite{raj2023targeted}. To achieve success in this case, the robot must react early and back up or move around the pedestrian, or else it will collide with the pedestrian.
    
    \textbf{(C4)}
    \textbf{Dynamic Obstacle Crossing the Path:}
    \label{subsection:c4}
    While C3 checks the situation when the dynamical obstacle moves directly towards the robot, here we test if the robot can react well when a pedestrian crosses the robot's straight line path perpendicularly. Fig.~\ref{fig:c4} shows this test case. This is similar to the \textit{Intersection} case in Raj et al.~\cite{raj2023targeted}. In this situation, even if the robot observes the pedestrian on its local map, it may not react in time as the obstacle is not yet on the global path. A successful execution requires the robot to back up to turn away from the pedestrian before moving ahead.
    
We compare the hybrid planner with DWA and SACPlanner across all these situations for 10 runs for C1, C2, and C3, and for 3 runs for C4. In C1 and C2, we also switch the start and goal location for half of the runs. As we focus on task efficiency, we compare the average distance traversed, velocity, time taken to navigate, and the number of collisions (in percentage) for each planner. 

\section{Results}
\label{sec:results}

\begin{figure*}[!ht]
    \vspace{3mm}
	\centering
	\subfigure{\label{fig:case1_hybrid}
		\includegraphics[width=0.24\textwidth]{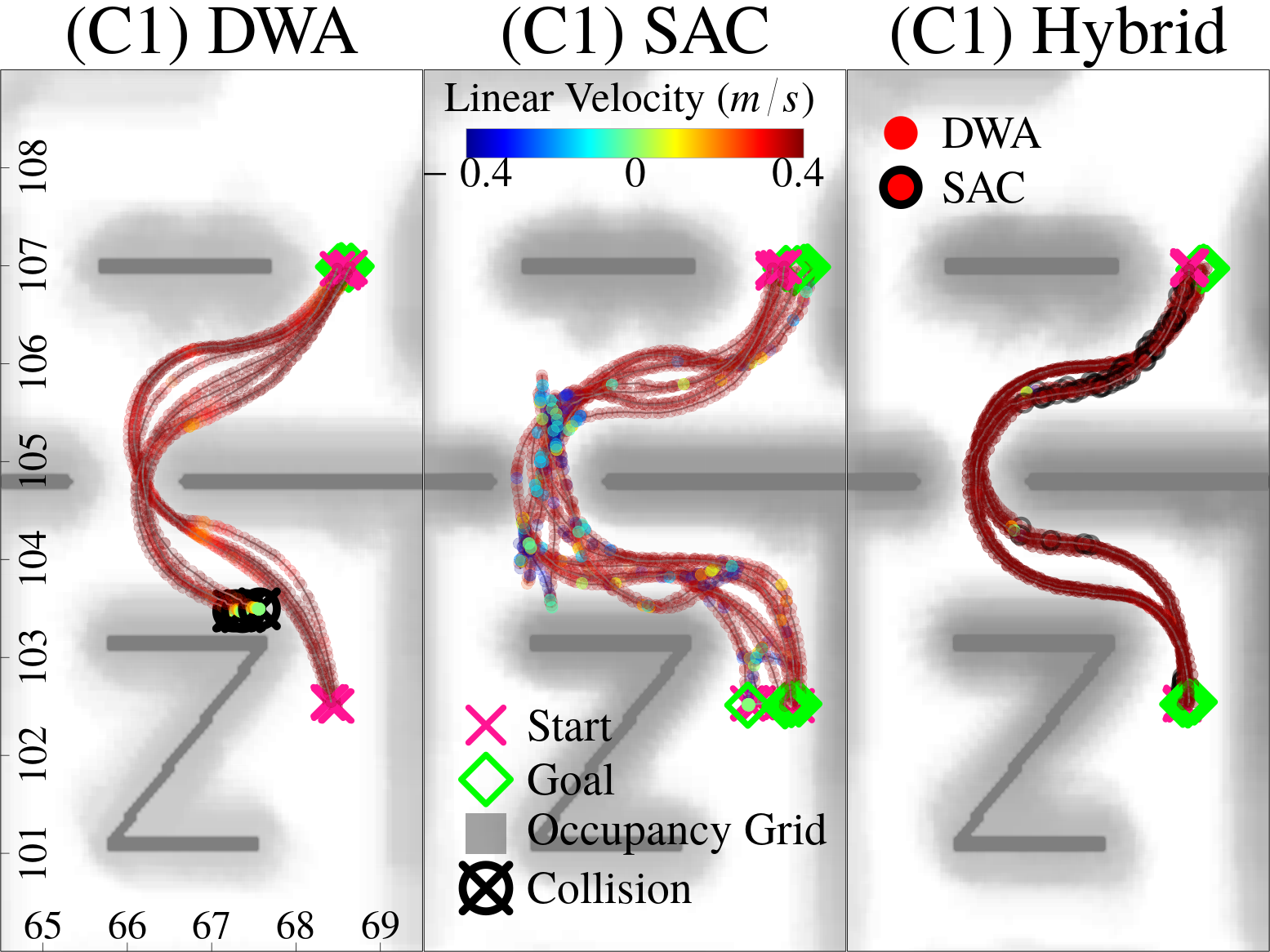}}\hspace{-1mm}
	\subfigure{\label{fig:case2_hybrid}
		\includegraphics[width=0.24\textwidth]{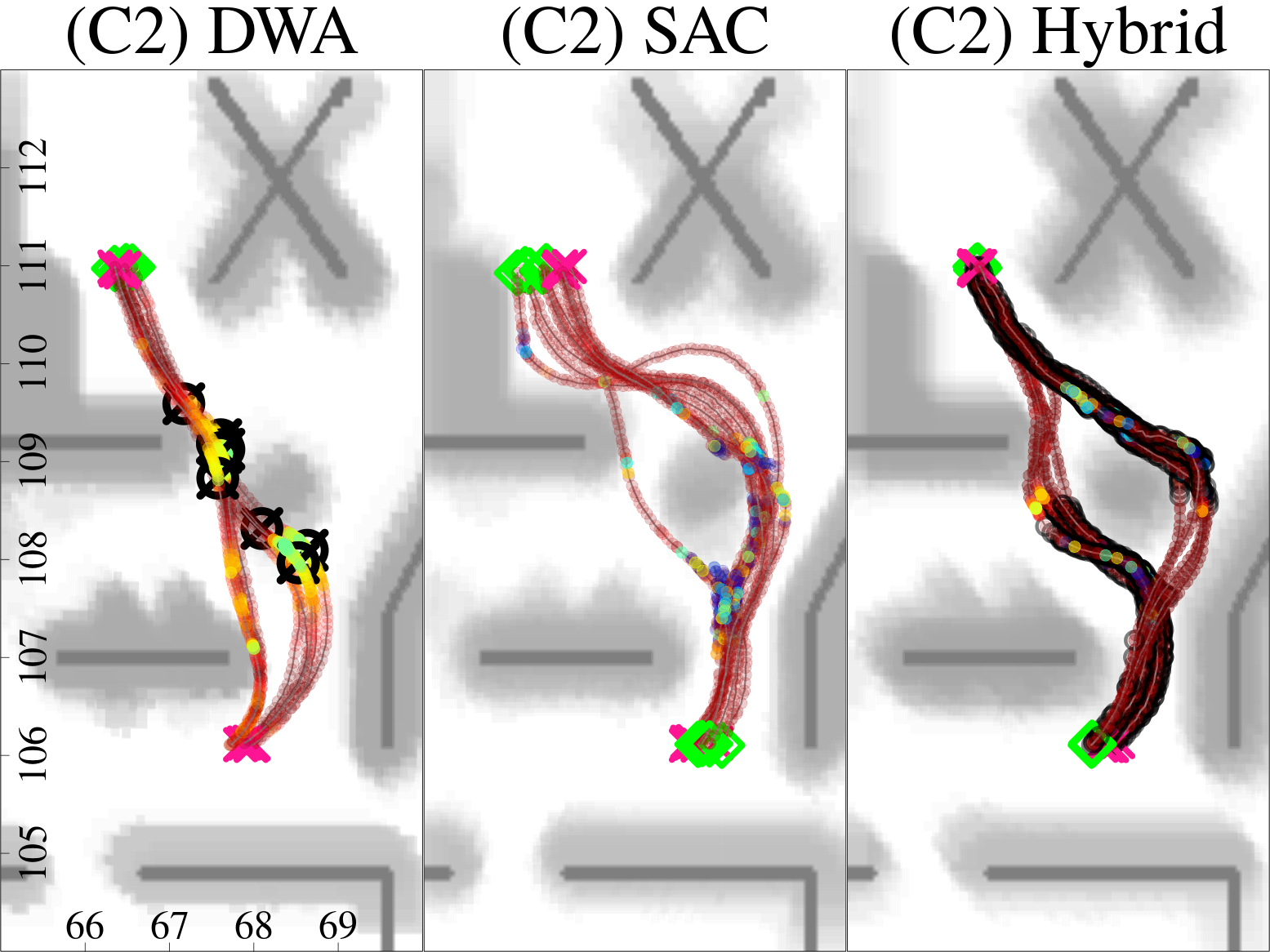}}\hspace{-1mm}	
	\subfigure{\label{fig:case3_hybrid}
		\includegraphics[width=0.24\textwidth]{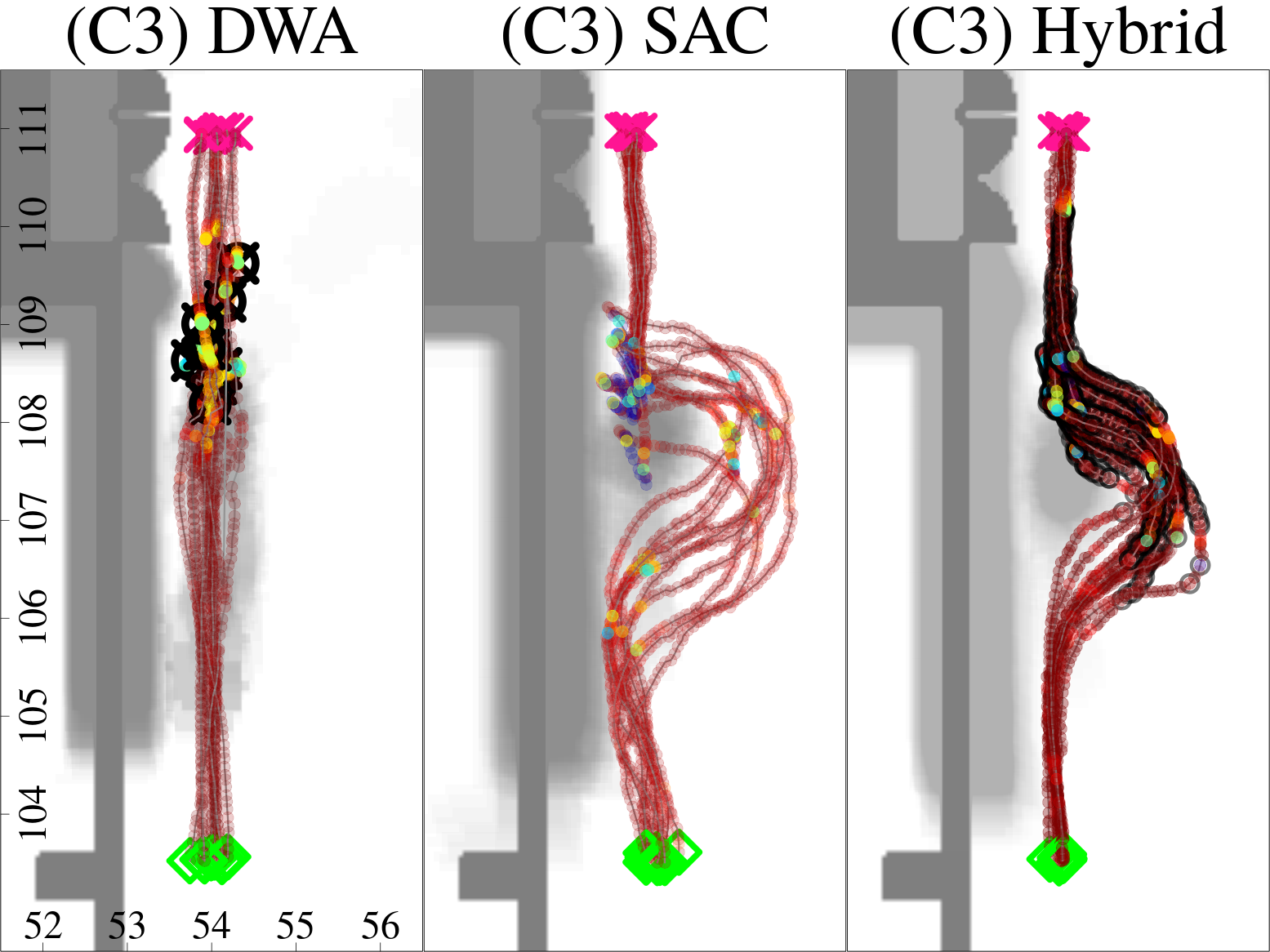}}\hspace{-1mm}	
	\subfigure{\label{fig:case4_hybrid}
		\includegraphics[width=0.24\textwidth]{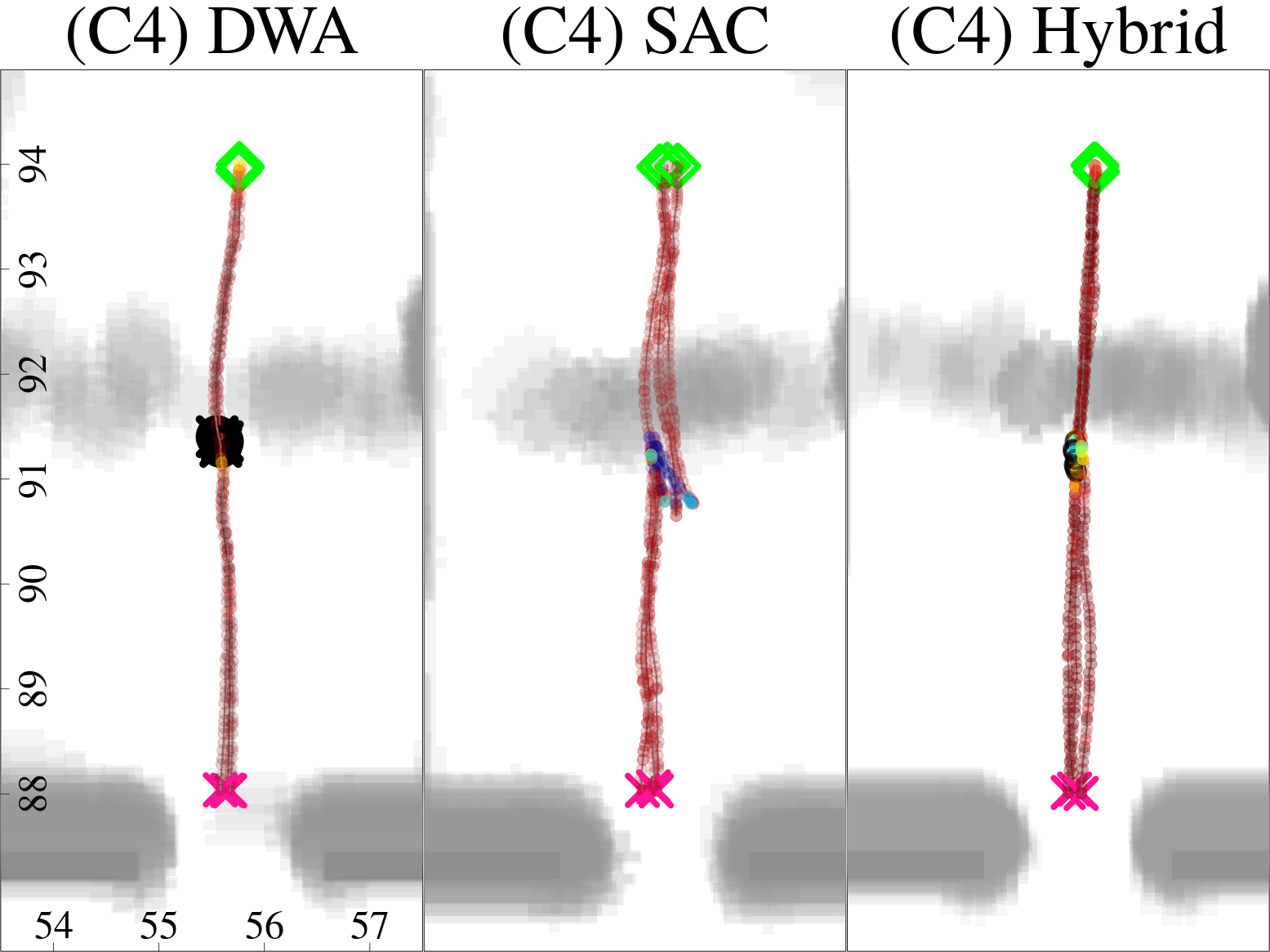}}	
	\vspace{-2mm}
	\caption{Trajectory comparison between DWA, SACPlanner vs. Hybrid planner agent for each test case.}\label{fig:case_hybrid}	
\end{figure*}

\begin{figure*}[!th]
	\centering
	\subfigure{\label{fig:ped_traj}
	\includegraphics[width=0.24\textwidth]{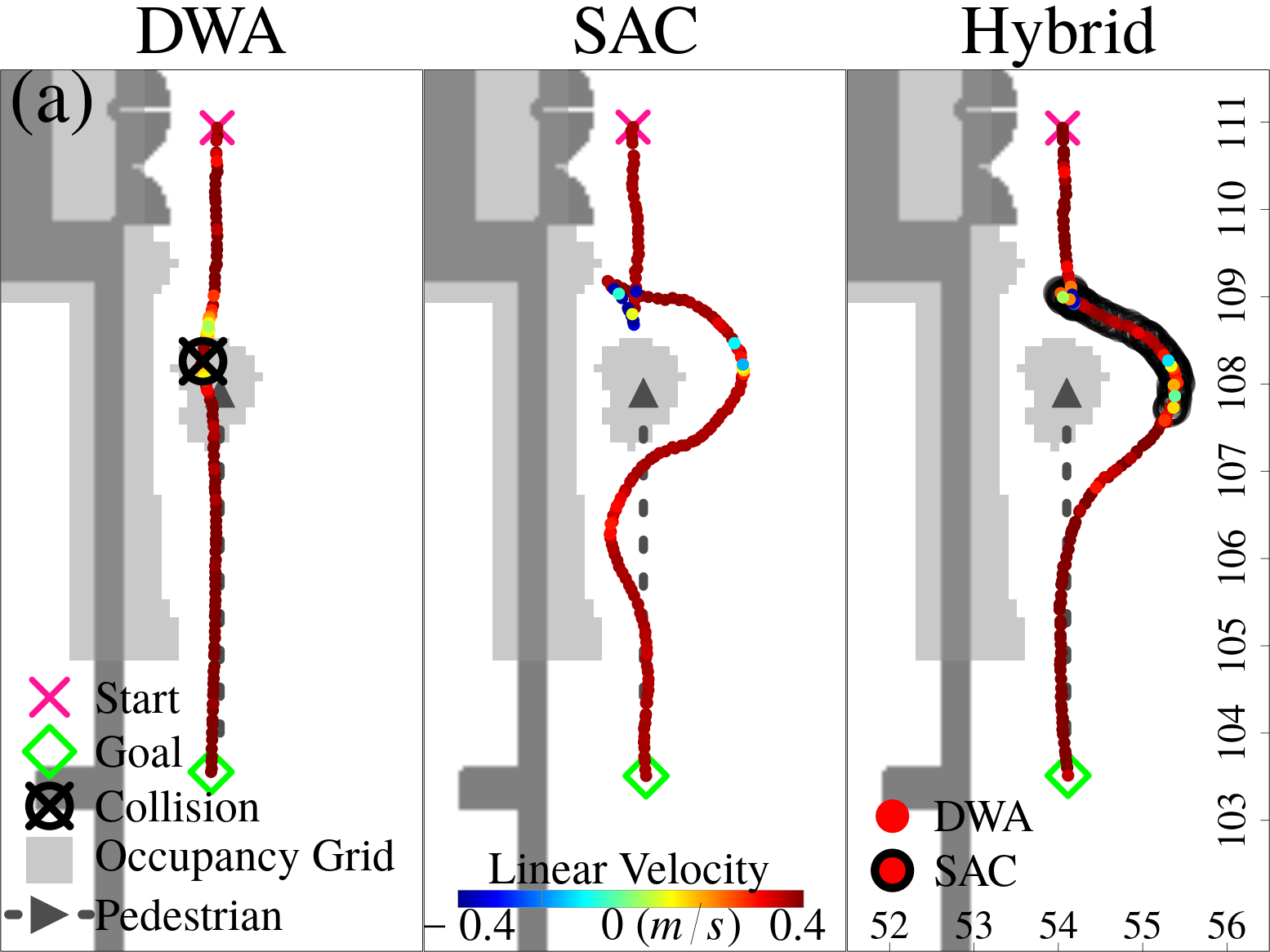}}	\hspace{-2mm}
	\subfigure{\label{fig:ped_linv}
	\includegraphics[width=0.24\textwidth]{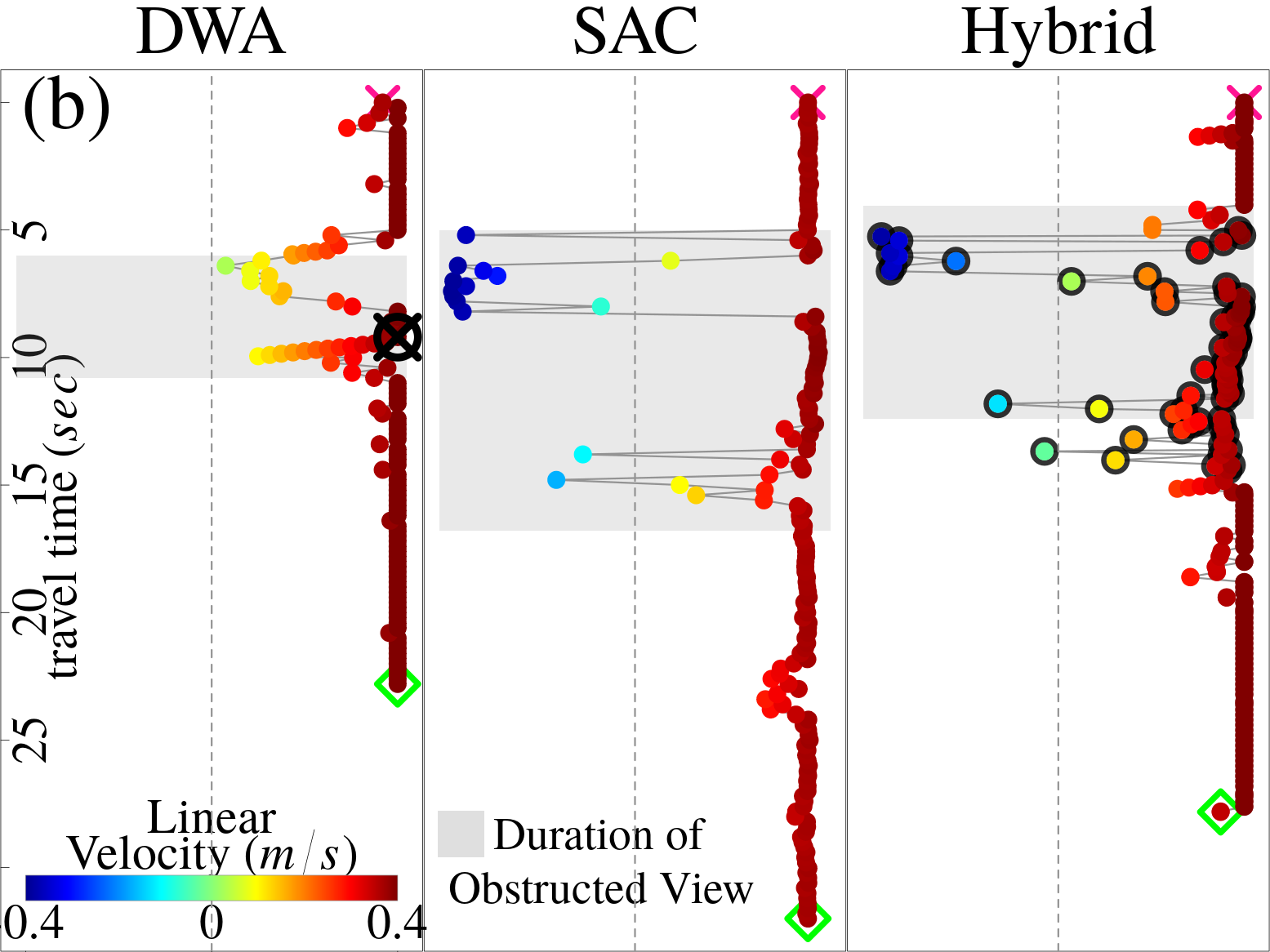}}	\hspace{-2mm}
	\subfigure{\label{fig:ped_angv}
	\includegraphics[width=0.24\textwidth]{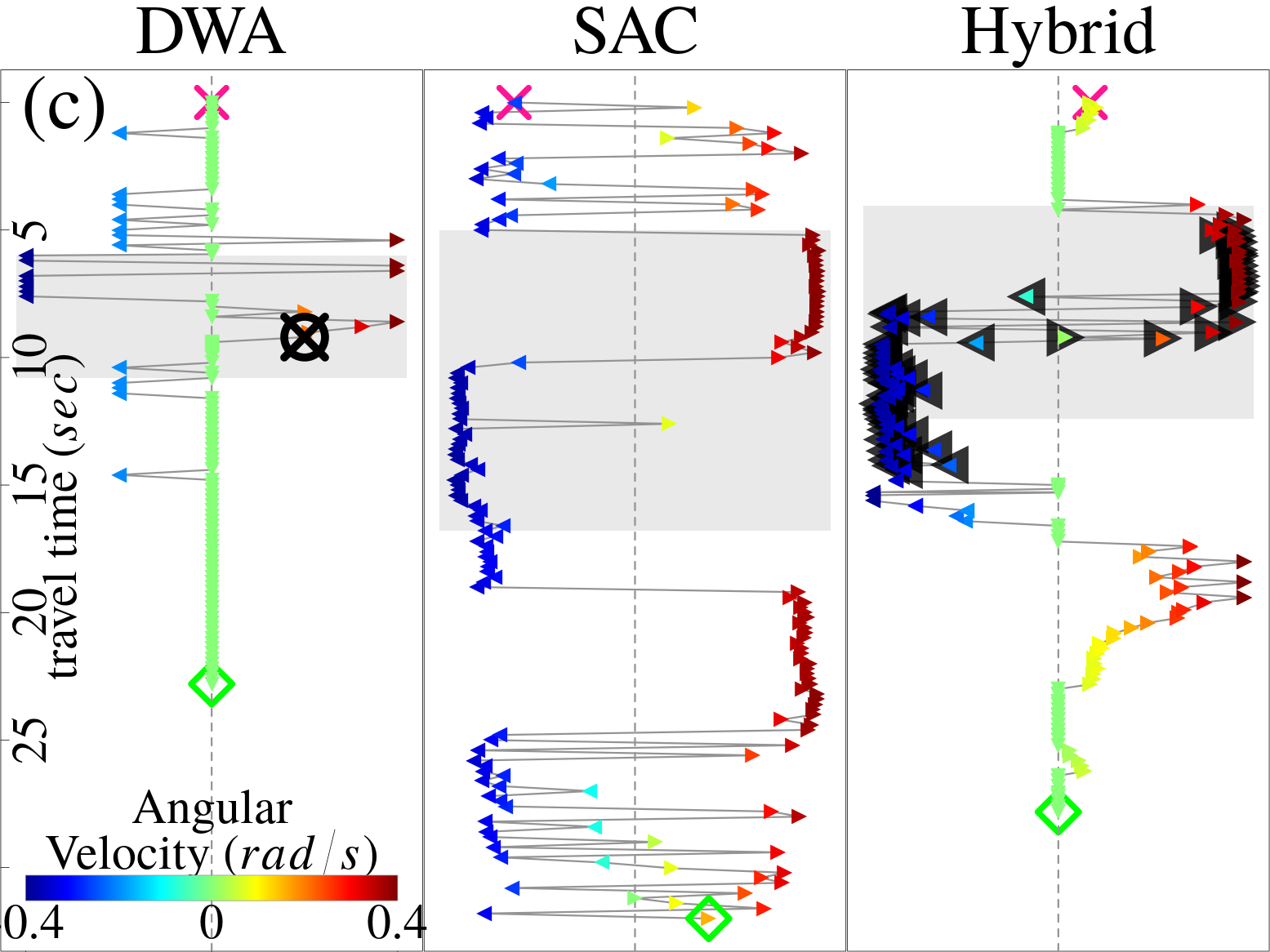}}\hspace{-2mm}
	\subfigure{\label{fig:ped_mdist}
	\includegraphics[width=0.24\textwidth]{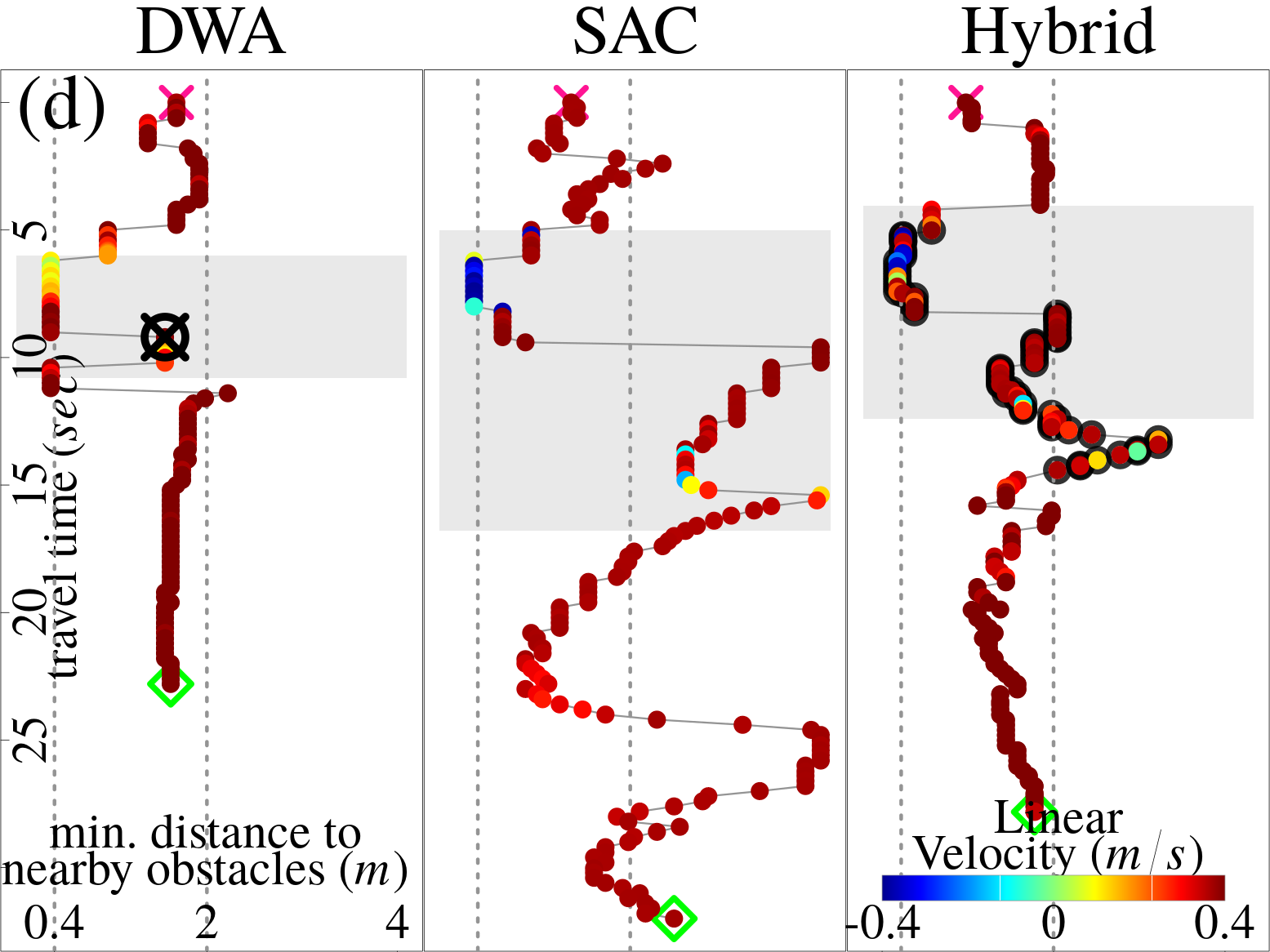}}
\vspace{-2mm}
	\caption{Trajectory comparison between DWA, SAC and Hybrid planners based on logs from the scenario (C3).}\label{fig:ped_ex}
\end{figure*}

The robot trajectories for each of C1-C4 are shown in Fig.~\ref{fig:case_hybrid}. We denote the start and goal along with the collision points. 
The color of the trajectory represents linear velocity and the circles with a thick black border represent where SACPlanner has been used for the hybrid planner. We also show the Occupancy Grid values in gray (taken from the map and the LiDAR). For C3 \& C4 the gray shading captures all the positions of the unexpected obstacle over time. The three local planners have qualitatively different behavior. DWA collides with the walls or obstacles in all cases except a few in C1. SACPlanner allows the robot to circumnavigate the obstacles but it results in the robot moving slowly, even with negative velocity in some cases, and usually results in a long detour. 
In each case, the hybrid planner helps the robot avoid obstacles successfully, while moving on a smooth trajectory with high speed, making it more suitable than either individual planner.

Table~\ref{tab:exp_results} summarizes the quantitative comparison averaged over 10 runs for C1-C3. (For brevity we refer to SACPlanner as SACs in the table.) The hybrid planner is faster than both DWA and SACPlanner, as shown by the higher average speed. Collisions when DWA is used, result in the robot covering a shorter distance without success. SACPlanner has the same success rate as the hybrid planner, but the hybrid planner results in a relative improvement of \textbf{26\%} in the navigation time with \textbf{18\%} shorter path length. Notably, our planner exhibits safe and efficient navigation in situations similar to prior works~\cite{raj2023targeted}, without the need to learn when to switch with a neural network.

\begin{table*}[!h]
	\caption{Summary statistics of trajectories from test cases.}
	\label{tab:exp_results}
	\centering{\normalsize
	\begin{tabular}
	{l|r|r|r|r|r|r|r|r|r}
	&	\multicolumn{3}{c|}{(C1)}					&	\multicolumn{3}{c|}{(C2)}					&	\multicolumn{3}{c}{(C3)}					\\\hline
	&	\multicolumn{1}{c|}{DWA}	&	\multicolumn{1}{c|}{SAC}	&	\multicolumn{1}{c|}{Hybrid}	&	\multicolumn{1}{c|}{DWA}	&	\multicolumn{1}{c|}{SAC}	&	\multicolumn{1}{c|}{Hybrid}	&	\multicolumn{1}{c|}{DWA}	&	\multicolumn{1}{c|}{SAC}	&	\multicolumn{1}{c}{Hybrid}	\\\hline
Time	&	21.80	&	37.20	&	{\bf 21.10}	&	30.70	&	28.50	&	{\bf 23.60}	&	27.50	&	33.10	&	{\bf 27.10}	\\\hline
Distance	&	{\bf 7.13}	&	10.70	&	7.67	&	{\bf 5.47}	&	8.57	&	7.41	&	{\bf 8.77}	&	10.80	&	9.40	\\\hline
Speed	&	0.33	&	0.29	&	{\bf 0.39}	&	0.18	&	0.30	&	{\bf 0.31}	&	0.32	&	0.33	&	{\bf 0.36}	\\\hline
Collision	&	50\%	&	0\%	&	{\bf 0\%}	&	100\%	&	0\%	&	{\bf 0\%}	&	100\%	&	0\%	&	{\bf 0\%}	\\\hline%

	\end{tabular}}
\end{table*}

To understand more deeply why the hybrid planner performs better, we show in Fig.~\ref{fig:ped_ex} the behavior of each planner in a single run from the test case (C3). The beginning and ending behavior of the hybrid planner is closer to a straight line, since DWA is selected using full-speed (dark red) linear velocities as in Fig.~\ref{fig:ped_traj}, \ref{fig:ped_linv}. The shaded area in Fig.~\ref{fig:ped_angv}-\ref{fig:ped_mdist} represents the duration of time when LiDAR first captures the pedestrian in its view in the polar costmap until he stops walking at the location $x=54$m, $y=108$m. From the overall travel time, the hybrid planner gets the robot to the goal faster than SACPlanner without any collisions. The reaction time (in seconds) to begin turning starting from when the robot first enters the shaded area, 4.05s for hybrid planner, 5s for SACPlanner and 5.99s for DWA planner. In addition, the hybrid planner gets around the pedestrian about 3.5 seconds faster than SACPlanner (8.36s $<$ 11.8s). The transition in rotational velocities is much smoother in the hybrid case since it reverts to DWA after passing around the pedestrian as in Fig.~\ref{fig:ped_angv}. Moreover, when the robot is far from the pedestrian the angular velocity is zero (green). This explains how the hybrid planner almost eliminates the jerky motion caused by SACPlanner. Fig.~\ref{fig:ped_mdist} shows the distance to the nearest `front obstacle' (within $\pm\frac{\pi}{4}$rad range from the current yaw). The hybrid planner manages both safe and efficient distance during the whole travel time.  

The results highlight that the hybrid planner makes appropriate use of both planners for navigation in various scenarios. It moves smoothly and quickly through clear areas and is responsive in face of obstacles discovered along the path. This behavior is also safer, both for the robot and for the humans acting as the dynamic obstacles.

\section{Discussion and Future Work}
\label{sec:conclusion}
We present a hybrid local planner that combines DWA, a classical planning method, and SACPlanner, a learning-based planning approach. Experiments on a ClearPath Jackal robot in various situations show that the proposed approach is safer and more efficient than the two constituent planners, showing a significant improvement in navigation time without any collision. The design of our switch forgoes the need to collect data and train another neural network, making it more suitable than learning-based switching from real-world development.

We focus on a heuristics-based approach to define the criteria for switching between the planners. Future work will explore more sophisticated approaches. A drawback of the hybrid approach is that the shortcomings of the constituent planners appear when the hybrid planner uses them. An example of this would be some jerky motion of the robot, owing to the SACPlanner, while the robot tried to avoid the obstacle. In the future, we intend to work on improving the constituent planners to further improve the overall performance of the hybrid planner.

\bibliographystyle{IEEEtran}
\bibliography{bibfile}

\end{document}